\documentclass[11pt]{article}

\usepackage[utf8]{inputenc}
\usepackage[T1]{fontenc}
\usepackage[margin=1in]{geometry}
\usepackage{newtxtext,newtxmath}
\usepackage{microtype}
\usepackage{graphicx}
\usepackage{booktabs}
\usepackage{natbib}
\usepackage[hidelinks]{hyperref}
\usepackage{xcolor}

\newcommand{\merit}{\textsc{MERIT}}

\title{When Does Memory Help? A Cost-Aware Evaluation of\\
Long-Term Memory in Tool-Using LLM Agents}
\author{Shweta Mishra\\
Independent Research\\
\texttt{1196shweta@gmail.com}
\and
Shashank Mishra\\
Independent Research\\
\texttt{24shashankm@gmail.com}}
\date{July 2026 (preprint v1.1)\thanks{All numbers in \S5 are measured
results: a two-generation pilot (9{,}940 scored episodes,
\texttt{gpt-4.1-mini}), run first with starter memory implementations and then real ones
(embedding retrieval, LLM summarization, LLM extraction) on the identical
grid, followed by the preregistered full grid (\S\ref{sec:phasec}):
3 agent models $\times$ 3 domains $\times$ 3 difficulty tiers with 3 seeds
on the primary model, 13{,}500 further episodes. Pilot and full grid:
23{,}440 scored episodes, \$42.57 in API cost; latest-generation and
frontier spot-checks (Claude Sonnet 5, Opus 4.8) are reported separately
as diagnostics in \S\ref{sec:threats}.
Code, traces, and preregistration: \url{https://github.com/smshweta/merit-bench}.}}

\begin{document}
\maketitle

\begin{abstract}
Long-term memory for LLM agents is evaluated today by conversational
recall benchmarks (LoCoMo, LongMemEval), which measure question answering
over dialogue history, not whether remembered facts change what a
\emph{tool-using} agent does. We present \merit{} (Memory Evaluation for
Realistic Instrumented Tasks), a benchmark and harness that measures the
\emph{marginal utility} of memory for task-executing agents under explicit
cost accounting. \merit{} provides episodic tool-use tasks in three
domains whose dependence on earlier-episode facts is verified by an
automated leak check; a difficulty ladder ending in \emph{updated-fact}
recall; controlled memory corruption; and full token and dollar metering
of every memory operation. Across 23{,}440 scored episodes (\$42.57), a
two-generation pilot on \texttt{gpt-4.1-mini} and a preregistered 3-model
$\times$ 3-seed grid (GPT-4.1, Claude Haiku 4.5; memory side held fixed),
memory lifts dependent-task success from a leak-verified floor of 0.00 to
0.55--1.00. On updated facts, embedding retrieval collapses
\emph{unpredictably} (0.30--0.95 across models; max seed gap 0.45), and
agents act on a correctly retrieved value only 55\% of the time, while
update-on-write stores (a structured fact store and, notably, LLM
summarization) remain at 0.70--1.00; the hybrid is worse than the fact
store alone. A latest-generation spot-check (Claude Sonnet 5, gated on a
clean full-replay control) reproduces the pattern. Swapping a memory's
implementation moves task success by up to 60 points, and full replay is
never economical: the best condition per domain delivers
2.7--3.9$\times$ its marginal utility per dollar. We release the
benchmark, harness, and all traces.
\end{abstract}

\section{Introduction}

LLM-based agents that plan, call tools, and act over multiple sessions are
moving rapidly into production: customer support, IT operations, coding, and
personal assistance. A central architectural question for every such
deployment is whether and how to give the agent \emph{long-term memory}:
information persisted across sessions and selectively recalled. A vibrant
ecosystem of memory systems has emerged (Mem0, Letta/MemGPT-style OS
memories, Zep, LangGraph checkpointing, bespoke RAG stores), and vendors
compete on recall benchmarks.

Yet the evaluation of agent memory has three blind spots:

\paragraph{Blind spot 1: Conversational QA is not task execution.}
The dominant benchmarks (LoCoMo \citep{locomo}, LongMemEval \citep{longmemeval},
and BEAM \citep{beam}) test whether a system can answer questions about
long, multi-session \emph{conversations}. Production agents, by contrast,
must use remembered facts to \emph{choose the correct tool call}: the right
customer ID, the previously agreed refund amount, the configuration fix
identified during last week's incident. Whether conversational recall scores
transfer to task-execution utility is an open empirical question.

\paragraph{Blind spot 2: Memory is assumed to help; harm is unmeasured.}
Recent work notes that agents sometimes ignore retrieved memory records or
use them inconsistently, and HaluMem \citep{halumem} examines memory
hallucination and consistency, but in conversational settings. In task
execution, a stale memory (a customer's \emph{old} address, a rolled-back
configuration) is not just unhelpful; it produces a \emph{confidently wrong
action}. No existing benchmark quantifies this failure mode under controlled
corruption, nor the more basic failure our pilot surfaces: agents that
demonstrably hold the correct fact in context and still do not act on it
(Ignore Rate 0.45--0.53 across implementations).

\paragraph{Blind spot 3: Cost is reported, but marginal utility is not.}
Memory systems report token counts alongside accuracy, but the decision
practitioners face is economic: does adding memory raise task success enough
to justify its per-task cost? We are not aware of any evaluation that reports
\emph{cost-adjusted marginal utility} (the change in success probability per
marginal dollar) across memory architectures.

We address all three with \merit{}. Our contributions:

\begin{enumerate}
\item \textbf{A task suite with controllable memory dependency and
difficulty.} Episodic tool-use tasks in three domains (customer support, IT
operations, personal assistant) where a tunable fraction of episodes depend
on facts established in earlier episodes, and a three-tier difficulty ladder:
\emph{easy} (one fact), \emph{medium} (multiple facts must be composed),
\emph{hard} (the fact is updated mid-arc and the latest value is required).
An automated \textbf{leak check} verifies at generation time that no gold
fact is present in the probe episode's inputs or re-derivable from tools.
\item \textbf{Controlled memory corruption.} Stale, contradictory, and
distractor records injected at known rates $\rho$, with ground-truth flags,
to measure Stale-Memory Harm.
\item \textbf{Three metrics absent from prior evaluations}: Memory
Utilization Rate (MUR), Ignore Rate, and Cost-Adjusted Marginal Utility
(CAMU), computed automatically by the harness (\S\ref{sec:metrics}).
\item \textbf{A two-generation pilot study} of 6 memory conditions $\times$
3 domains $\times$ 3 difficulty tiers (9{,}940 scored episodes; the full
grid run twice, with starter implementations and with real ones) with
preregistered hypotheses and statistics (paired bootstrap clustered by arc;
Holm--Bonferroni), demonstrating that the benchmark separates memory
architectures that are indistinguishable under single-fact recall, and
separates \emph{architecture} from \emph{implementation quality}; plus
a 3-agent-model $\times$ 3-seed replication (13{,}500 further episodes,
\S\ref{sec:phasec}) showing the updated-fact collapse of embedding
retrieval is model- and seed-unstable while update-on-write memories are
uniformly robust.
\item \textbf{Open-source release} of the benchmark, harness, and all
traces, with a deterministic \$0 mock-model mode that validates the full
pipeline.
\end{enumerate}

\section{Related Work}

\subsection{Memory systems for LLM agents}

MemGPT \citep{memgpt} frames memory as an OS-style hierarchy with the LLM
paging information in and out of context. Mem0 \citep{mem0} extracts and
consolidates structured facts at write time; Zep \citep{zep} maintains a
temporal knowledge graph; A-Mem \citep{amem} organizes agentic notes
dynamically. Rolling summarization is the default pattern in agent frameworks
such as LangChain/LangGraph. These systems papers evaluate primarily on
conversational QA (LoCoMo, LongMemEval); none evaluate task-execution utility
under cost accounting. \merit{}'s conditions C1--C5 are deliberately
\emph{reimplementations of these architecture families} under one interface,
so that architectural mechanisms (retrieval, summarization, update-on-write,
hybrid) can be compared in isolation from product engineering. We
deliberately evaluate reimplementations rather than the shipped products:
a product's score conflates its architectural mechanism with proprietary
retrieval tuning, prompting, and engineering, preventing causal
attribution to the mechanism; benchmarking shipped systems on \merit{} is
a natural follow-up enabled by the released harness.

\subsection{Benchmarks for agent memory}

LoCoMo \citep{locomo} evaluates QA over very long multi-session dialogues.
LongMemEval \citep{longmemeval} covers information extraction, multi-session
reasoning, temporal reasoning, knowledge updates, and abstention over
${\sim}$115K-token histories. BEAM \citep{beam} scales conversational probing
to 10M tokens. HaluMem \citep{halumem} evaluates hallucination at the level
of memory operations (extraction, updating, QA). RealMem \citep{realmem}
moves toward project-oriented interaction. All are \emph{answer-producing}
evaluations; \merit{} is \emph{action-producing}: success is a predicate over
the final state of a mutable world, and the knowledge-update dimension that
LongMemEval and HaluMem probe conversationally becomes, in \merit{}'s hard
tier, a behavioral test of whether the agent \emph{acts} on the latest value.
Long-context benchmarks (RULER, \citealp{ruler}; BABILong, \citealp{babilong})
test single-pass attention over a given context, not the write-and-retrieve
loop of a persistent memory.

\subsection{Tool-use and agent benchmarks}

$\tau$-bench \citep{taubench} evaluates tool-calling agents conversing with
simulated users over domain APIs with database-state success checks (the
closest environment design to \merit{}'s), but each task is a single episode;
memory across episodes is not exercised. AgentBench \citep{agentbench},
SWE-bench \citep{swebench}, and WebArena \citep{webarena} similarly evaluate
within-episode competence. \merit{} adds the cross-episode dependency
structure (arcs, plants, probes, leak check) on top of a $\tau$-bench-style
environment.

\subsection{Reliability of agents}

Production reports consistently rank reliability as the top barrier to agent
deployment; confidently-wrong actions are costlier than abstentions.
Stale-memory harm is a concrete, measurable instance: \merit{} injects
staleness at known rates and measures both the success drop and whether the
agent re-verifies before acting.

\section{The \merit{} Benchmark}

\subsection{Design goals}

G1 (Ecological validity): tasks require tool calls with programmatically
verifiable end states. G2 (Controllable dependency and difficulty): the
dependent-task ratio and the difficulty tier are benchmark parameters. G3
(Adversarial realism): corruption reflects real staleness processes. G4 (Cost
transparency): every token in and out of the memory system is metered. G5
(Reproducibility): deterministic seeded worlds, scripted or cached simulated
users, pinned model versions, released traces, and a \$0 deterministic mock
model that exercises the entire pipeline.

\subsection{Environment}

An \textbf{episode} is one agent session with a simulated user and a tool API
over a mutable world state (SQLite). Episodes are grouped into \textbf{arcs}
of 4--6 episodes sharing entities. Arcs contain \textbf{plant} episodes (a
fact is established conversationally; the user explicitly forbids acting on
it yet), optional \textbf{update} episodes (the fact is revised),
\textbf{probe} episodes (success requires the fact; it is absent from the
episode's inputs), and \textbf{independent} episodes (solvable
within-episode).

Two safeguards proved essential in practice. The \textbf{leak check} asserts
at generation time that every gold fact value is absent from the probe
episode's inputs and from the initial world state, with gold values kept
unique arc-wide. \textbf{Delta scoring} marks an episode successful only if
its checker predicate flips from false to true \emph{during} that episode:
in early gate runs, eager agents processed refunds during plant episodes, and
probes then ``succeeded'' off inherited world state; delta scoring eliminates
this world-state leak channel entirely (across 9{,}940 episodes, the 22 probes
that arrived with their checker already satisfied score as failures, never as
inherited successes).

\subsection{Domains}

\begin{itemize}
\item \textbf{D1 Customer support (retail):} tools = \texttt{get\_order},
\texttt{refund}, \texttt{update\_address}, \texttt{get\_policy},
\texttt{send\_message}. Facts: agreed partial-refund amounts, updated
addresses.
\item \textbf{D2 IT operations:} tools = \texttt{search\_logs},
\texttt{get\_deploy\_history}, \texttt{get\_config}, \texttt{set\_config},
\texttt{deploy}, \texttt{update\_ticket}. Facts: diagnosed config fixes, safe
rollback versions.
\item \textbf{D3 Personal assistant:} tools = \texttt{get\_calendar},
\texttt{create\_event}, \texttt{send\_email}, \texttt{get\_preference},
\texttt{set\_preference}. Facts: standing room/time preferences, dinner
commitments (place, day, time).
\end{itemize}

\subsection{Difficulty ladder and memory conditions}

\textbf{Difficulty}: \emph{easy} = one gold fact per probe; \emph{medium} =
multi-fact probes (every gold value must appear in the executed tool calls);
\emph{hard} = updated-fact probes: the value is planted, revised in a later
episode, and the probe requires the \textbf{latest} value. The hard tier is a
targeted stressor for the architectural distinction between stores that
overwrite (update-on-write) and stores that accumulate (replay, retrieval).

\textbf{Memory conditions}, one interface (\texttt{write(episode)},
\texttt{read(context)}): C0 none; C1 full replay of prior transcripts; C2
retrieval (starter: keyword-overlap; real: embedding retrieval,
\texttt{text-embedding-3-small}); C3 rolling summary (starter: extractive
truncation; real: LLM summarization); C4 structured fact store with
update-on-write (starter: pattern-based extraction; real: LLM extraction);
C5 hybrid (C4 + C2). Each condition ran in both an inexpensive
\textbf{starter} implementation and a \textbf{real} one (memory-side
LLM/embedding calls, fully metered); the grid was run once per generation
(\S\ref{sec:swap} compares them). All conditions share the same agent
scaffold (ReAct-style tool loop; \citealp{react}), prompts (except the
memory block), tools, and decoding (temperature 0).

\subsection{Metrics}\label{sec:metrics}

\begin{itemize}
\item \textbf{TSR}: task success rate by programmatic end-state check, split
by dependent/independent.
\item \textbf{MUR}: among dependent episodes where all gold values were
present in the retrieved memory block, the fraction where every gold value
appears in the executed tool-call arguments (value tracing). A frozen,
stratified 100-episode audit sample is committed to the repository. A
human annotator labeled all 100 items blind to the tracer's output: the
tracer agrees with human judgment on 93\% of items (Cohen's $\kappa =
0.63$, substantial agreement). All seven disagreements are in the same
direction, with the tracer scoring \emph{not utilized} where the human
judged the value was used (typically reformatted values the containment
check misses), so the tracer under-counts utilization and the Ignore
Rates we report are conservative upper bounds. Independent
second-annotator labeling is left to future work.
\item \textbf{Ignore Rate} $= 1 - \mathrm{MUR}$ on episodes with correct
memory present.
\item \textbf{SMH}: $\mathrm{TSR}(\text{clean}) -
\mathrm{TSR}(\text{corrupted})$ at $\rho \in \{0.1, 0.3\}$ for stale /
contradiction / distractor corruption.
\item \textbf{Cost \& CAMU}: metered tokens and dollars per episode;
$\mathrm{CAMU} = \Delta\mathrm{TSR}$ vs.\ C0 per $\Delta$cost vs.\ C0;
break-even task value $= \Delta\text{cost}/\Delta\mathrm{TSR}$.
\end{itemize}

\subsection{Statistical methodology}

Paired comparisons on identical task instances; 95\% CIs and two-sided $p$ by
paired bootstrap (10{,}000 resamples) clustered at the arc level;
Holm--Bonferroni within each preregistered hypothesis family (H1--H4,
committed to the public repository before experiments; commit history is the
preregistration record).

\section{Experimental Setup}

Model: \texttt{gpt-4.1-mini} (pinned; temperature 0) on both the agent side
and, in the real-implementation generation, the memory side (embeddings:
\texttt{text-embedding-3-small}). Scale per generation: 10 arcs $\times$ 5
episodes per (domain $\times$ difficulty $\times$ condition); dependent-task
ratio 0.5; corruption sweep (3 modes $\times$ $\rho \in \{0.1, 0.3\}$) with
LLM-paraphrased users on D1. Totals: \textbf{9{,}940 scored episodes, \$9.12
in API cost}; starter generation 5{,}440 episodes / \$4.61, real
generation 4{,}500 episodes / \$4.52 (${\sim}\$0.001$/episode including
memory-side calls, which are metered into all cost figures). Harness:
Python, LiteLLM for provider-agnostic calls and metering; SQLite worlds;
scripted simulated users (LLM paraphrase mode with a paraphrase-leak check
on D1). The mock-model mode replays the entire grid deterministically at \$0
and is exercised by 62 unit tests, including generation determinism, leak
checks, and end-to-end pipeline invariants. Unless marked otherwise,
\S5.1--5.6 report the real-implementation generation on
\texttt{gpt-4.1-mini}.

\paragraph{Full grid (\S\ref{sec:phasec}).} The preregistered multi-model
grid runs the real implementations on three agent models:
\texttt{gpt-4.1-mini} (3 seeds), GPT-4.1 (frontier, 1 seed), and Claude
Haiku 4.5 (cross-vendor, pinned \texttt{claude-haiku-4-5-20251001}, 1
seed), over the identical 3-domain $\times$ 3-tier grid, 10 arcs
$\times$ 5 episodes per cell. The memory side is pinned to
\texttt{gpt-4.1-mini} (embeddings: \texttt{text-embedding-3-small}) for
\emph{every} agent model, so the agent model is the only varying factor.
Totals: \textbf{13{,}500 scored episodes, \$33.45} (\$20.46 OpenAI +
\$12.99 Anthropic), bringing the study to 23{,}440 episodes and \$42.57
overall.

\section{Results}

\subsection{Memory helps on dependent tasks; the floor is real}

C0 scored \textbf{0.000} on dependent tasks in all nine (domain $\times$
difficulty) cells, in both generations: the leak check holds; there is no
route to the gold facts except memory. On independent tasks C0 scores
0.83--1.00, confirming task solvability. Every memory condition beats C0 on
dependent tasks in every domain at easy difficulty ($\Delta$TSR $+0.55$ to
$+1.00$; all Holm-adjusted $p \le 0.001$, paired bootstrap clustered by
arc).

\subsection{The difficulty ladder dissociates architectures}\label{sec:ladder}

\begin{table}[t]
\centering
\caption{Dependent-task TSR by difficulty tier and memory condition, real
implementations (D1~/~D2~/~D3). See also Figure~\ref{fig:ladder}.}
\label{tab:ladder}
\small
\begin{tabular}{lccccc}
\toprule
Tier & C1 replay & C2 retrieval & C3 summary & C4 facts & C5 hybrid \\
\midrule
easy   & 1.00 / 1.00 / 0.90 & 0.90 / 1.00 / 0.90 & 0.95 / 1.00 / 0.70 & 1.00 / 0.55 / 0.80 & 0.95 / 0.80 / 1.00 \\
medium & 0.60 / 1.00 / 0.95 & 0.30 / 0.95 / 1.00 & 0.30 / 1.00 / 0.90 & 0.80 / 0.40 / 0.75 & 0.80 / 1.00 / 0.95 \\
hard   & 1.00 / 0.95 / 1.00 & 0.70 / 0.35 / 0.45 & 1.00 / 0.70 / 1.00 & 0.95 / 0.75 / 1.00 & 0.80 / 0.60 / 0.50 \\
\bottomrule
\end{tabular}
\end{table}

\begin{figure}[t]
\centering
\includegraphics[width=\textwidth]{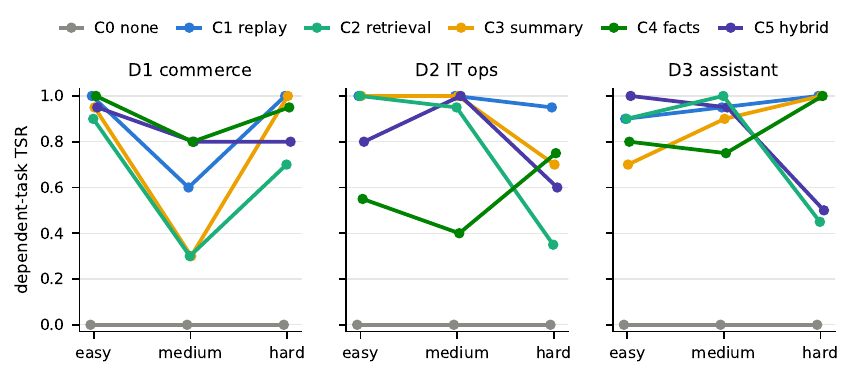}
\caption{Dependent-task TSR across the difficulty ladder, per memory
condition and domain (real implementations). The hard (updated-fact) tier
separates memories that overwrite state, i.e., the fact store (C4) and, notably,
LLM summarization (C3), which rewrites its summary each episode, and
chronological replay (C1) from retrieval-based memories (C2, C5).}
\label{fig:ladder}
\end{figure}

Table~\ref{tab:ladder} and Figure~\ref{fig:ladder} show three dissociations:
(1)~\textbf{Updated facts break retrieval memory, including embedding
retrieval}: C2 falls to 0.35--0.70 on hard while C1 stays at 0.95--1.00 and
C4 at 0.75--1.00: retrieval surfaces stale and fresh values side by side
with no recency signal, while update-on-write overwrites and replay resolves
recency from chronology. (2)~\textbf{Hybrid is worse than its better half}
on hard in all three domains (0.50--0.80 vs.\ C4's 0.75--1.00): the
retrieval half re-imports the staleness the fact store had eliminated.
(3)~\textbf{Full replay is update-immune but composition-limited}: C1 drops
to 0.60 on D1-medium despite containing every fact: possessing information
and composing it are different capabilities. A fourth pattern emerged only
with real implementations: \textbf{LLM summarization behaves like
update-on-write} (1.00 / 0.70 / 1.00 on hard), because regenerating the
summary each episode naturally keeps the latest value, an architectural
kinship invisible in the starter generation (\S\ref{sec:swap}).

\subsection{Starter vs real implementations: the swap is diagnostic}
\label{sec:swap}

\begin{figure}[t]
\centering
\includegraphics[width=\textwidth]{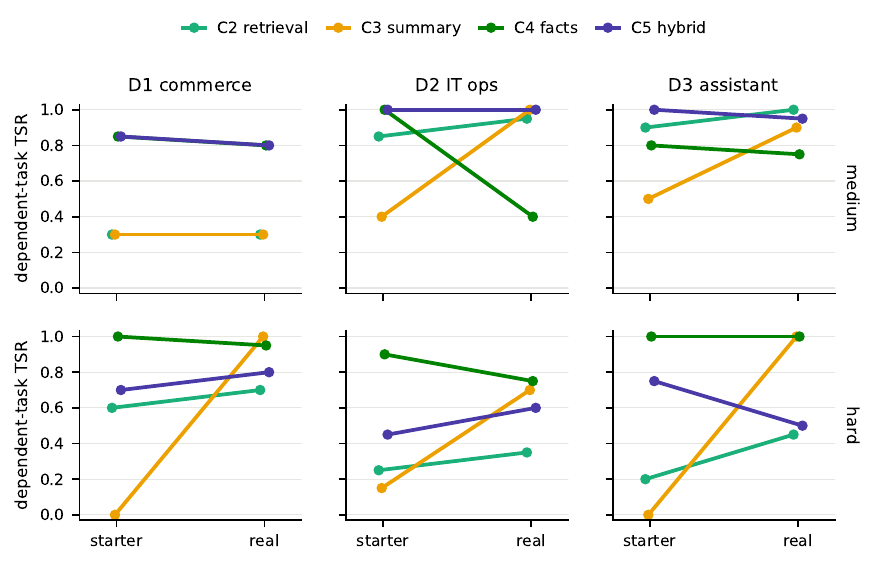}
\caption{Dependent-task TSR, starter vs real implementation of each memory
condition (C2--C5), on the medium (top) and hard (bottom) tiers. LLM
summarization rescues C3 on updated facts; LLM extraction costs C4 up to 60
points in D2; embedding retrieval does not fix C2's hard-tier collapse.}
\label{fig:swap}
\end{figure}

Upgrading C2--C5 in place, on the identical grid, moves conditions in both
directions (Figure~\ref{fig:swap}):

\begin{itemize}
\item \textbf{C3 summary, truncation $\to$ LLM:} 0.00 / 0.15 / 0.00 $\to$
\textbf{1.00 / 0.70 / 1.00} on hard. The starter result was a floor set by
the implementation, not the architecture: truncation drops facts (starter
MUR $= 1.0$ on the episodes it retained), while an LLM summarizer that
rewrites state each episode is update-robust by construction.
\item \textbf{C4 facts, patterns $\to$ LLM extraction:} D2 falls from 1.00
to \textbf{0.40} on medium (0.60 $\to$ 0.55 easy, 0.90 $\to$ 0.75 hard)
while D1/D3 roughly hold. The hand-tuned patterns the starter needed per
domain (\S\ref{sec:discussion}) were not dead weight; the generic LLM
extractor misses IT-ops facts they caught. Extraction quality is a real tax
on structured memory, payable in either engineering effort or metered tokens
and now visible in the benchmark.
\item \textbf{C2 retrieval, keyword $\to$ embedding:} modest gains
everywhere ($+0.10$ to $+0.25$ on hard), but the hard-tier collapse persists
(the failure is the architecture's missing recency arbitration, not
retrieval quality).
\end{itemize}

The two generations bound an implementation-sensitivity band per
architecture; the band is wide (up to 1.00 TSR for C3 on updated facts),
which is itself an argument for action-level evaluation over memory-system
benchmarking.

\subsection{Agents ignore memories they hold}

\begin{figure}[t]
\centering
\includegraphics[width=\textwidth]{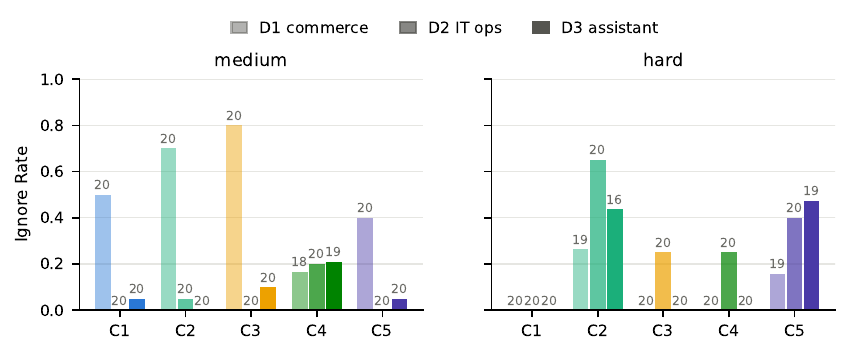}
\caption{Ignore Rate (the fraction of dependent episodes where every gold
value was present in the memory block but the agent did not act on it) by
condition and domain on the medium and hard tiers (real implementations).
Numbers above bars are episode counts with memory present (bars at zero are
shown by their count only).}
\label{fig:ignore}
\end{figure}

On the hard tier, pooling domains, the correct (latest) value was present in
C2's retrieved block in 55 probe episodes; the agent acted on it in 30
(\textbf{Ignore Rate 0.45}; 0.53 in the starter generation; upgrading
retrieval quality barely moves it). Even C1, whose memory block is a clean
full transcript, ignores up to 0.50 of held facts in multi-fact
(medium-tier) episodes (Figure~\ref{fig:ignore}). This is Blind spot~2 made
measurable: memory-system accuracy overstates end-task benefit unless
utilization is measured.

\subsection{Stale-memory harm}\label{sec:smh}

\begin{figure}[t]
\centering
\includegraphics[width=\textwidth]{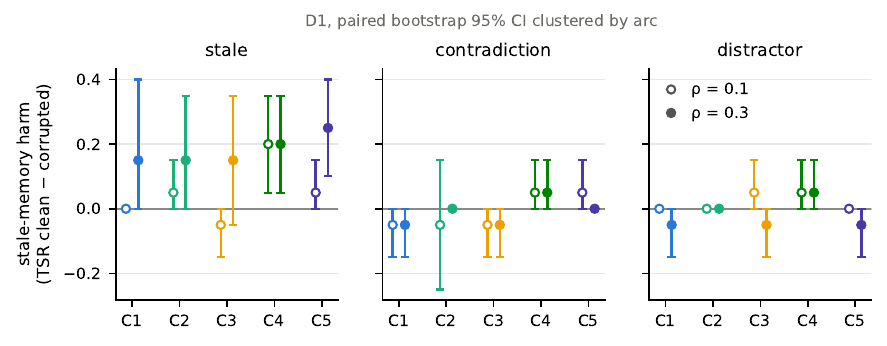}
\caption{Stale-memory harm (TSR clean $-$ corrupted, dependent tasks, D1,
real implementations) by condition, corruption mode, and corruption rate
$\rho$; error bars are paired bootstrap 95\% CIs clustered by arc.}
\label{fig:smh}
\end{figure}

On D1 with stale corruption (Figure~\ref{fig:smh}), the largest and only
Holm-significant harm is again on the hybrid C5 (SMH $+0.25$ at $\rho =
0.3$, Holm-adjusted $p = 0.030$), replicating the starter-generation
result with real implementations. One shift is instructive: C4 with LLM
extraction shows stale harm of $+0.20$ (raw $p = 0.013$, not surviving Holm)
where regex-C4's harm was ${\le}0.05$: the LLM extractor ingests corrupted
records that the rigid patterns rejected, another face of the
extraction-quality tax (\S\ref{sec:swap}). Contradiction and distractor
corruption produce small, non-significant effects at pilot scale.
(Corruption sweeps beyond D1, with verification-rate analysis, are left to
future work.)

\subsection{Cost and marginal utility}

\begin{figure}[t]
\centering
\includegraphics[width=\textwidth]{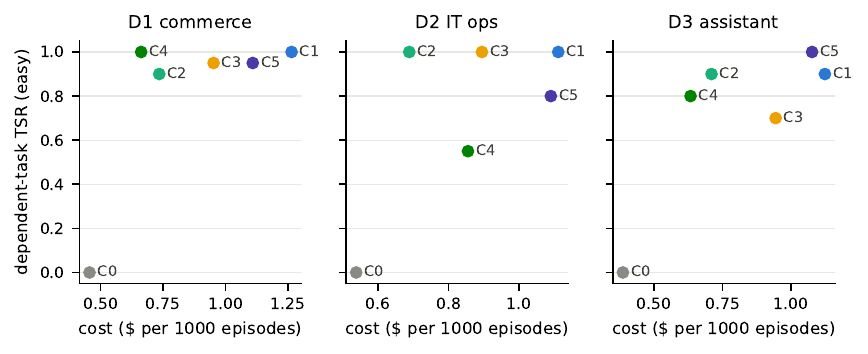}
\caption{Metered cost per episode (including memory-side calls) vs.\
dependent-task TSR at the easy tier, real implementations. C4 stays on the
Pareto frontier in D1/D3 but cedes it in D2; C1 full replay pays a
2--3$\times$ cost premium for equal or lower TSR everywhere.}
\label{fig:cost}
\end{figure}

With memory-side calls metered, per-episode cost on D1-easy
(Figure~\ref{fig:cost}): C0 \$0.00046, C4 \$0.00066, C2 \$0.00073, C3
\$0.00095, C5 \$0.00111, C1 \$0.00126 (2{,}914 tokens/episode;
$2.7\times$ C0's cost). CAMU and accuracy rankings differ in every domain
(H4 supported): the best CAMU is C4 in D1 (4{,}839 points of dependent-task
success per marginal dollar) and D3 (3{,}245), but \textbf{C2 in D2}
(6{,}629), where LLM extraction's accuracy regression (\S\ref{sec:swap})
erased C4's edge. Full replay is never the economical choice
(1{,}222--1{,}741 pts/\$; 2.7--3.9$\times$ worse than the per-domain best).
Two starter-generation artifacts vanish under honest metering: C4's headline
${\sim}17{,}300$ pts/\$ (regex extraction was free) becomes 4{,}839, and its
\emph{negative} marginal cost in D2 becomes $+\$0.0003$: the extraction
tax, now on the books. Break-even task values remain fractions of a cent for
all conditions: memory pays for itself at trivially low task values
\emph{when it works}; the practitioner-relevant differences are in
robustness (\S\ref{sec:ladder}--\ref{sec:smh}), not raw affordability, at
these model prices. The absolute CAMU figures are tied to mid-2026 API
pricing and will drift as prices change; the qualitative result, that
robustness rather than per-episode cost is the deciding factor and that
full replay never leads on CAMU, is price-independent, since it rests on
the ratio of success gains to cost \emph{differences} between conditions.

\subsection{The full grid: seed robustness and cross-model generality}
\label{sec:phasec}

The full grid (\S4) asks two questions the pilot could not: do the pilot's
dissociations survive resampling (3 seeds on \texttt{gpt-4.1-mini}), and do
they survive a change of agent model (GPT-4.1; Claude Haiku 4.5, a
different vendor and tier) with the memory side held fixed?

\paragraph{Seed variance concentrates in embedding retrieval.} Across the
nine 3-seed cells, the maximum pairwise TSR gap between seeds averages
0.16 for C2, the largest of any condition, with a worst cell of 0.45
(C2, D1-hard: per-seed TSR 0.45 / 0.90 / 0.75). C1 averages 0.07; the
overwrite-style memories C3 and C4 average 0.11--0.12. A single-seed
evaluation of embedding retrieval on the hard tier could honestly report
anywhere from 0.45 to 0.90 for the same system; seed-mean numbers with
spread are the only defensible summary, and all numbers below are seed
means.

\begin{figure}[t]
\centering
\includegraphics[width=\textwidth]{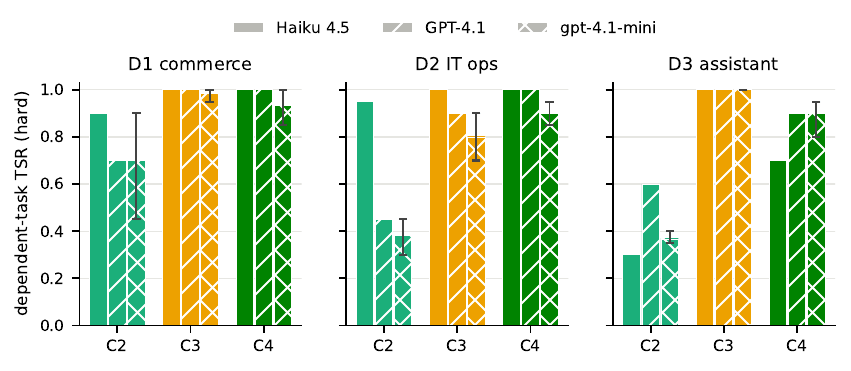}
\caption{Hard-tier TSR across agent models (bar hatch = model, color =
condition; whiskers = min--max across the 3 \texttt{gpt-4.1-mini} seeds).
Embedding retrieval (C2) collapses somewhere for every model, but where is
model-idiosyncratic; LLM summarization (C3) is robust everywhere.}
\label{fig:phasec}
\end{figure}

\paragraph{The hard-tier collapse replicates across models: a
reliability failure, not a fixed deficit.} Table~\ref{tab:phasec} and
Figure~\ref{fig:phasec} report the hard (updated-fact) tier for all three
models. Embedding retrieval (C2) is the only architecture that collapses,
but \emph{where} it collapses is model-idiosyncratic: Haiku 4.5 holds the
single-fact domains (0.90 / 0.95 on D1 / D2) where GPT-4.1 and
\texttt{gpt-4.1-mini} drop to 0.38--0.70, yet Haiku falls hardest on
multi-fact D3 (0.30). No model escapes: every model has at least one
domain at ${\le}0.60$, and the spread across model $\times$ domain is
0.30--0.95. LLM summarization (C3), by contrast, spans 0.80--1.00 on the
same cells (robust for every model, domain, and seed), and the
structured fact store (C4) spans 0.70--1.00.

\begin{table}[t]
\centering
\small
\begin{tabular}{llccc}
\toprule
hard tier, TSR\textsubscript{dep} & & Haiku 4.5 & GPT-4.1 & gpt-4.1-mini ($\pm$sd) \\
\midrule
C2 embedding retrieval & D1 & 0.90 & 0.70 & 0.70 $\pm$ 0.23 \\
 & D2 & 0.95 & 0.45 & 0.38 $\pm$ 0.08 \\
 & D3 & 0.30 & 0.60 & 0.37 $\pm$ 0.03 \\
\midrule
C3 LLM summarization & D1 & 1.00 & 1.00 & 0.98 $\pm$ 0.03 \\
 & D2 & 1.00 & 0.90 & 0.80 $\pm$ 0.10 \\
 & D3 & 1.00 & 1.00 & 1.00 $\pm$ 0.00 \\
\midrule
C4 structured fact store & D1 & 1.00 & 1.00 & 0.93 $\pm$ 0.08 \\
 & D2 & 1.00 & 1.00 & 0.90 $\pm$ 0.05 \\
 & D3 & 0.70 & 0.90 & 0.90 $\pm$ 0.09 \\
\bottomrule
\end{tabular}
\caption{Hard (updated-fact) tier across agent models, memory side pinned
to \texttt{gpt-4.1-mini}. C0 floor is 0.00--0.05 and C1 full replay is
1.00 for every model (omitted). Full per-cell tables for all tiers are in
the released traces.}
\label{tab:phasec}
\end{table}

The dissociation is significant for every model. Pooling domains on the
hard tier, the paired arc-clustered bootstrap gives C3$-$C2 $= +0.44$
$[+0.36, +0.53]$ for \texttt{gpt-4.1-mini}, $+0.38$ $[+0.23, +0.53]$ for
GPT-4.1, and $+0.28$ $[+0.15, +0.43]$ for Haiku 4.5 (all $p \le 0.0002$;
C4$-$C2 positive for all, $p \le 0.009$), and $+0.25$ $[+0.10, +0.40]$
($p = 0.002$) for the Sonnet 5 spot-check cell. The per-domain contrasts
localize Haiku's entire deficit in D3 ($+0.70$ $[+0.45, +0.95]$; D1/D2
not significant): the significance structure itself traces the
model-idiosyncratic locus of collapse.

The refined claim is stronger than the pilot's: on updated facts,
embedding retrieval is not merely worse: it is \textbf{unpredictable},
both across seeds (max gap 0.45) and across agent models (0.30--0.95),
because success depends on whether that particular agent resolves the
stale-versus-fresh conflict in retrieved context, an ability that varies
non-monotonically with model tier. Update-on-write architectures remove
the conflict at write time and are correspondingly stable everywhere. A
practitioner choosing a memory system from a single-model, single-seed
benchmark number would systematically over- or under-estimate embedding
retrieval; MERIT's grid makes the variance itself measurable.

\paragraph{A latest-generation check.} As a final probe we ran the D1-hard
cell on Claude Sonnet 5, a 2026-generation agentic model, gated on the same
C1 control that disqualified Opus 4.8 (\S\ref{sec:threats}): here the control is clean (C1 $=$ 1.00, C0 floor
0.00). The pattern replicates: embedding retrieval (C2) drops to 0.75 while
LLM summarization (C3) and the fact store (C4) hold 1.00 (hybrid C5 0.90).
The updated-fact weakness of retrieval-only memory is not an artifact of
older model generations.

Each real implementation remains one representative of its family (one
embedding model, one summarization prompt, one extraction prompt), and
\S\ref{sec:swap} shows how much such choices matter.

\section{Discussion}\label{sec:discussion}

\paragraph{Practitioner guidance (provisional).}
If tasks depend on facts that get \emph{revised} (addresses, configs,
schedules; that is, most operational facts), prefer a memory that \emph{overwrites
state} (a structured fact store, or LLM summarization, which turns out to
be update-robust because it rewrites its summary every episode) over
retrieval, which accumulates. Do not assume a hybrid inherits the better
component's behavior: measure it. Full replay is a strong accuracy
baseline that fails on cost ($2.7\times$ tokens) and on multi-fact
composition. And treat the write path as a first-class risk: swapping
extraction implementations moved C4 by 60 points in one domain
(\S\ref{sec:swap}); truncation is not summarization.

\paragraph{Why do agents ignore correct memories?}
Traces show two patterns: (i)~under retrieval, stale and fresh values
co-occur and the agent averages, asks, or picks the stale one (no
provenance/timestamps to arbitrate); (ii)~under multi-fact composition,
agents act on the subset of facts nearest the task phrasing. Both suggest
memory \emph{presentation} (provenance, recency marking, contradiction
surfacing) matters as much as memory \emph{storage}.

\paragraph{Extraction brittleness as a hidden cost, now measured.}
In the starter generation, C4's pattern extractor was blind outside its home
domain until hand-tuned patterns were added; the real generation shows the
converse: generic LLM extraction pays for its generality with a 60-point
accuracy regression in D2 and greater willingness to ingest stale records
(\S5.5). In production, extraction quality is the tax that structured memory
pays, payable in domain engineering or in metered tokens, but never zero.
\merit{} puts it on the books (CAMU includes memory-side calls).

\paragraph{A methodological note.}
Two of our safeguards were added because early runs failed without them:
eager agents leaked gold facts into world state (fixed by delta scoring), and
gold-value collisions leaked across task pairs (fixed by arc-wide
uniqueness). Task-execution memory benchmarks have leak channels that
conversational QA benchmarks cannot express; we recommend leak checks over
\emph{world state}, not just prompts, as standard practice.

\section{Limitations and Threats to Validity}\label{sec:threats}

\textbf{Construct:} programmatic checkers may not capture all real-world
success notions; the MUR tracer is string containment, validated against a
single human annotator ($\kappa = 0.63$, conservative direction) but not
yet by independent double annotation; the human
audit (the frozen sample is committed). \textbf{Internal:} prompt
differences across conditions are confined to the memory block; delta
scoring removes world-state carryover; the mock-model grid guards the
pipeline, but mock results are never reported as findings.
\textbf{External:} the grid spans three agent models plus two gated
spot-checks, and the spot-checks bound the model-strength question in both
directions: frontier capability does not substitute for memory (C0
$=$ 0.00 even for Claude Opus 4.8 and Sonnet 5), and the C2 collapse
persists on the newest generation. Yet all findings remain within three
synthetic domains with scripted users (LLM-paraphrase mode mitigates
phrasing overfit on D1), 10 arcs per cell, and each real memory
implementation is a single representative of its family;
\S\ref{sec:swap} quantifies how consequential implementation choices are. \textbf{Reproducibility:} deterministic seeded generation,
pinned model IDs, released traces, \$0 mock mode. API model deprecation
remains a limitation: the released traces preserve the reported runs, and
the harness accepts any OpenAI-compatible endpoint (including locally
served open-weight models), so the grid is rerunnable even after the
reported API models retire.

\paragraph{Frontier-model spot-check: task success conflates memory use
with policy prudence.} After the full grid we ran one diagnostic cell
(D1-hard, 6 conditions $\times$ 50 episodes) on Claude Opus 4.8, the
strongest model available to us. The no-memory floor held (C0 $=$ 0.00):
frontier capability cannot substitute for memory on leak-verified dependent
tasks. But the full-replay control collapsed to C1 $=$ 0.75 (vs 1.00 for
all three grid models), and the transcripts show why: on updated-fact
probes the model \emph{quotes the remembered amounts and then declines to
act on them}, observing that the update trail ``traces back only to my own
confirmation messages'' repeating user-asserted values, and requesting
supervisor confirmation for what it reads as a chat-escalated refund,
arguably correct behavior for a production agent, and plausibly a product
of safety training. Two implications. First, raw TSR comparisons across
models conflate memory use with policy prudence; the C1 control detects
exactly this confound, and cross-model numbers should be read relative to
each model's own C1. Second, \merit{}'s updated-fact arcs (a user
renegotiating an amount across sessions) are structurally similar to
social-engineering escalations, and safety-tuned models may increasingly
treat \emph{memory provenance} as part of the decision. Benchmarks that
score only task completion will under-credit such models; we report this
cell as a diagnostic, not a leaderboard entry.
By contrast, Claude Sonnet 5, the same vendor's current agentic-tuned
model, passes the control cleanly (C1 $=$ 1.00) and reproduces the
pattern of \S\ref{sec:phasec} (C2 0.75 vs C3/C4 1.00 on D1-hard): the
prudence confound is a property of specific safety postures, not of model
generation, and the C1 control suffices to detect it per model.

We flag this as more than a measurement nuisance. It suggests a distinct
axis that action-level memory evaluation must contend with as models are
increasingly safety-tuned: \emph{memory-provenance skepticism}, in which an
agent that correctly recalls a value nonetheless declines to act because it
cannot establish that the value was legitimately authorized rather than
user-asserted. On \merit{}'s updated-fact arcs this manifests as a lower
ceiling on TSR that is not a memory failure at all (the fact is retrieved
and quoted) and that a naive leaderboard would misread as a weaker agent.
Whether the behavior is desirable (it plausibly prevents social-engineering
in production) or over-cautious (it blocks legitimate task completion) is
context-dependent and beyond our scope; our claim is narrower: the C1
full-replay control isolates it from genuine memory failure, so cross-model
TSR should always be read relative to each model's own C1. This observation
rests on a single diagnostic cell (one domain, 50 episodes) and two
frontier models; characterizing provenance skepticism across domains,
prompts, and model families is left to future work, but \merit{}'s control
structure already makes it visible and separable, which we believe is a
prerequisite for studying it at all.

\section{Conclusion}

\merit{} reframes agent-memory evaluation from ``can the system recall?'' to
``does recall change what the agent does, at what cost, and how does it
fail?''. The answer is not monotone: architectures indistinguishable on
single-fact recall separate by up to 0.70 TSR when facts must be
superseded (C3 1.00 vs.\ C2 0.30, Haiku 4.5, D3-hard), agents ignore
nearly half of correctly retrieved
facts, swapping one memory implementation for another moves task success by
as much as 1.00 TSR in either direction, and the most accurate memory is
2.7--3.9$\times$ less economical than the most efficient one. The benchmark,
harness, traces, and preregistration are public; we invite memory-system
authors to evaluate against \merit{}.

\end{document}